\documentclass[lettersize,journal]{IEEEtran}
\usepackage{amsmath,amsfonts}
\usepackage{algorithm}
\usepackage{algpseudocode}
\usepackage{array}
\usepackage{textcomp}
\usepackage{stfloats}
\usepackage{url}
\usepackage{verbatim}
\usepackage{graphicx}
\usepackage{cite}
\usepackage{color}
\usepackage{bbm}  
\usepackage{subfigure}

\usepackage{booktabs} 
\usepackage{multirow} 
\usepackage{ulem}

\usepackage{xcolor}  

\usepackage{lineno}

\usepackage{hyperref}
\hypersetup{
	colorlinks=true,
	linkcolor=blue,
	citecolor=blue,
	urlcolor=blue
}

\usepackage{amsmath, amssymb, amsfonts}
\usepackage{mathtools}      
\usepackage{bm}             
\usepackage{siunitx}        
\usepackage{booktabs}       
\usepackage{algorithm}
\usepackage{algpseudocode}  

\usepackage{tabularx}        
\usepackage{array}           
\usepackage{multirow}        
\usepackage{makecell} 
\usepackage{ragged2e}   

\newcolumntype{Y}{>{\centering\arraybackslash}X}
\newcolumntype{P}[1]{>{\raggedright\arraybackslash}p{#1}}

\newcommand{\best}[1]{\textbf{#1}}
\newcommand{\secondbest}[1]{\uline{#1}}

\begin{document}


\title{PRISM: Prioritized Channel Importance with Semi-supervised Domain Adaptation for Cross-Subject EEG Emotion Recognition}

\author{Xin Zhou, Xiang Zhang, Hao Deng, and Lijun Yin$^{*}$, \textit{Fellow, IEEE} 

\thanks{Xin Zhou, Xiang Zhang and Lijun Yin are with the School of Computing, T. J Watson College of Engineering and Applied Science, Binghamton University - State University of New York, Binghamton, NY 13902 USA (e-mail: xzhou11@binghamton.edu; zxiang4@binghamton.edu; lyin@binghamton.edu). 
}

\thanks{Hao Deng is with the Massachusetts General Hospital, Harvard University, Boston, MA 02114 USA (e-mail: hdeng1@mgh.harvard.edu).}

\thanks{$^*$Corresponding author.}
}



\maketitle

\begin{abstract}
Electroencephalogram (EEG) captures endogenous brain activity with high temporal fidelity and holds substantial promise for precise emotion decoding. However, channel redundancy and pronounced inter-subject variability remain key obstacles to scalable generalization. To address these limitations, we propose a novel framework termed \textbf{PR}ioritized channel \textbf{I}mportance with \textbf{S}emi-supervised do\textbf{M}ain adaptation (PRISM), enabling label-efficient cross-subject emotion decoding. On the channel side, PRISM assigns differentiable, data-dependent channel weights via a lightweight expert ensemble, amplifying reliable electrodes while suppressing distractors. On the domain side, PRISM leverages unlabeled data through confidence-filtered pseudo-labels to drive consistency regularization and domain alignment, mitigating subject-specific heterogeneity. Extensive experiments show that PRISM surpasses state-of-the-art methods on DEAP, DREAMER, and SEED datasets, achieving robust cross-subject generalization given limited annotations. 
\end{abstract}


\section{Introduction}
\IEEEPARstart{E}EG is noninvasive and has high temporal resolution, which enables the capture of affect related neural dynamics and is therefore regarded as an ideal signal for emotion decoding \cite{pan2022matt, kobler2022spd}. Neuropsychological studies indicate that emotion processing exhibits regional selectivity across the cortex, with frontal systems showing particular sensitivity \cite{coan2004frontal}. In practice, some electrodes contribute little to emotional representations and are more susceptible to ocular and myogenic artifacts \cite{gong2023astdf, li2024multi}, which leads to pronounced spatial nonuniformity in full channel EEG. Using all channels without discrimination dilutes discriminative information and reduces recognition accuracy, and it also increases dimensional redundancy and computational cost. Identifying and emphasizing electrodes that are more informative for emotion decoding, while suppressing redundant and noisy sources, is therefore a key path to improving the quality and deployability of EEG-based emotional representations.

Prior work has explored emotion recognition with a small set of channels and found that using only a limited number of emotion-relevant electrodes as input does not markedly reduce accuracy \cite{yang2025seeg, zhou2025mimamba}. Other studies employ attention mechanisms \cite{yang2025seeg, tao2020eeg} or graph convolutions \cite{lin2023eeg, yang2024automatically} to assign dynamic weights across channels. However, many existing approaches either do not adequately account for differences in cortical responses across distinct emotion elicitation paradigms, or they rely on a single weighting configuration, which limits adaptability across tasks, paradigms, and settings. Given heterogeneous elicitation conditions and application constraints, supporting multiple weighting configurations that update in a data adaptive manner is both practically meaningful and methodologically valuable.

Beyond channel redundancy, EEG exhibits pronounced cross-subject heterogeneity, that is, substantial innate differences among individuals in anatomy, physiological state, and psychological responses. As a result, the EEG distributions produced by different individuals under the same elicitation conditions can differ markedly, and even the same subject may drift over time \cite{zhou2024eegmatch}. These distributional discrepancies make the shift between source and target subjects one of the primary causes of degraded cross subject recognition performance. Techniques such as feature alignment \cite{zhu2025multi}, subdomain adaptation \cite{li2024gusa, ju2025domain}, and adversarial graph contrastive learning \cite{ye2024semi} have made progress in mitigating this issue. However, they often require many labels or highly accurate pseudo labels, and they seldom model intra EEG structure explicitly, for example, channel level differences, which leaves training sensitive to noise and to pseudo-label drift. To cope with label scarcity, these methods are often paired with semi-supervised \cite{zhou2024eegmatch, ye2024semi} and unsupervised learning strategies \cite{li2024gusa, zhang2023unsupervised, zhoubrainuicl}. However, they typically rely on additional auxiliary components such as graph neural networks or attention mechanisms, or they lack tight integration with standard backbones, which complicates practical use and limits plug-and-play deployment.

Building on the discussion above, we can summarize that EEG-based emotion recognition faces two main challenges:

\begin{itemize}
	\item Which EEG channels are most informative under different emotion elicitation conditions, and how can a model elevate electrodes that contribute to specific emotions while suppressing interference from redundant channels?
	
	\item How can cross-subject heterogeneity be mitigated, particularly in target settings with scarce labels, so that the learned representations remain reliable and generalizable?
\end{itemize}

To this end, we think that it is necessary to prioritize channel importance, and there is a pressing need for an end-to-end framework that, under label scarcity, simultaneously strengthens model generalization and performs domain alignment. Inspired by advances in mixture-of-experts (MoE) \cite{eigen2013learning} and semi-supervised domain adaptation \cite{berthelot2021adamatch}, we adopt multiple lightweight expert sub-networks that operate in parallel and select a subset of experts conditioned on the input and task, thereby instantiating multiple weighting configurations that naturally fit EEG channel prioritization. In addition, semi-supervised domain adaptation integrates supervised learning, unsupervised consistency regularization, and domain-alignment constraints, which directly addresses the cross-subject setting with limited labels.

Accordingly, we propose PRISM (PRioritized channel Importance with Semi-supervised doMain adaptation), a framework that, across diverse EEG emotion recognition tasks, assigns data-dependent soft weights to each channel and performs cross-subject, semi-supervised domain adaptation under limited labels. Specifically, PRISM first encodes spatiotemporal EEG features with a backbone network, then augments it with a lightweight expert ensemble that learns differentiable, adaptive per-channel weights to amplify reliable electrodes while suppressing distractors. In parallel, confidence-filtered pseudo labels on unlabeled target data support consistency regularization and domain alignment, which mitigates heterogeneity and improves generalization. The framework is model agnostic and compatible with mainstream time-series architectures, readily accommodating emotion recognition across different label densities.

The main contributions of this paper can be summarized
as follows:

\begin{itemize}
	
	\item We propose PRISM, which realizes channel prioritization via a lightweight expert ensemble, yielding learnable multi-weight configurations that adapt to diverse emotion-elicitation paradigms and task settings.
	
	\item Under label-scarce circumstances, we develop and validate a semi-supervised domain adaptation strategy tailored to EEG, significantly improving cross-subject robustness and label efficiency.
	
	\item On public benchmarks including DEAP, DREAMER, and SEED, PRISM consistently outperforms state-of-the-art methods under limited annotations, and it can be integrated in a plug-and-play manner into existing models to further enhance performance.
\end{itemize}

The remaining sections of this article are organized as follows. Section \ref{sec2} reviews the related work on channel selection, mixture of experts and semi-supervised learning. Section \ref{sec3} presents the pipeline of PRISM. Section \ref{sec4} details the procedure of the conducted experiments and experimental results. A more in-depth discussion is provided in Section \ref{sec5}. Finally, the study is concluded in Section \ref{sec6}.

\section{Related Work}\label{sec2}
\subsection{Channel Selection}
The brain engages distinct regions across cognitive activities \cite{ding2025emt}. Converging evidence indicates that the frontal and temporal lobes are associated with emotion \cite{yang2025seeg, tao2020eeg, yang2024automatically, gong2023eeg}, with particularly strong effects in frontal regions \cite{ding2025emt, guo2024convolutional}. Negative and neutral emotions show greater activation in the prefrontal cortex, whereas positive emotions are more active in the left hemisphere \cite{li2024multi}. Tao et al. \cite{tao2020eeg} introduced an attention mechanism to adaptively allocate weights and observed higher weights for electrodes over the frontal, temporal, and parietal areas. Lin et al. \cite{lin2023eeg} regulated the proportion of selected channels by leveraging attention distributions on a graph structure. Similarly, Yang et al. \cite{yang2024automatically} employed a channel weighting network to estimate channel importance parameters. Selecting channels that contribute more to emotion recognition does not reduce accuracy and can improve model interpretability \cite{yang2025seeg}.

\subsection{Mixture of Experts}
MoE \cite{eigen2013learning} instantiates multiple submodels and uses a gating network or router to dynamically select a small subset of experts for each input. It has been widely adopted in natural language processing, computer vision, and time series prediction. For example, Switch Transformers \cite{fedus2022switch} and GShard \cite{lepikhin2020gshard} maintain massive parameter counts while controlling compute, thereby improving efficiency. V-MoE \cite{riquelme2021scaling} routes capacity preferentially to target regions and downweights background. MMVAE \cite{shi2019variational} combines MoE to fuse latent representations from different modalities. Methods such as Pathformer \cite{chen2024pathformer}, Time-MoE \cite{shi2024time}, InterpGN \cite{wen2025shedding}, and SoftShape \cite{liu2025learning} assign different experts to different scales, which improves model stability and interpretability.

\subsection{Semi-supervised Learning}
Semi-supervised learning requires only a small number of labels while achieving strong target-domain generalization. Early work MixMatch \cite{berthelot2019mixmatch} combines label guessing, entropy minimization, consistency regularization, and MixUp \cite{zhang2017mixup} to form an efficient semi-supervised framework. FixMatch \cite{sohn2020fixmatch} uses high-confidence pseudo labels together with a constraint that enforces consistency between weak and strong augmentations, leading to strong performance. AdaMatch \cite{berthelot2021adamatch} provides a unified training framework that covers semi-supervised learning, unsupervised domain adaptation, and semi-supervised domain adaptation. FlexMatch \cite{zhang2021flexmatch} and FreeMatch \cite{wang2022freematch} adopt more flexible threshold selection strategies to adapt across classes. SoftMatch \cite{chen2023softmatch} replaces hard thresholds with Gaussian weighting. AllMatch \cite{wu2024allmatch} fully exploits unlabeled data through class-adaptive thresholds and class-consistency constraints. Similarly, FullMatch \cite{chen2023boosting} integrates FixMatch and FlexMatch and can also maximize the use of all unlabeled data.

\section{Method}\label{sec3}
\begin{figure*}[t]
	\centering
	\includegraphics[width=0.85\linewidth]{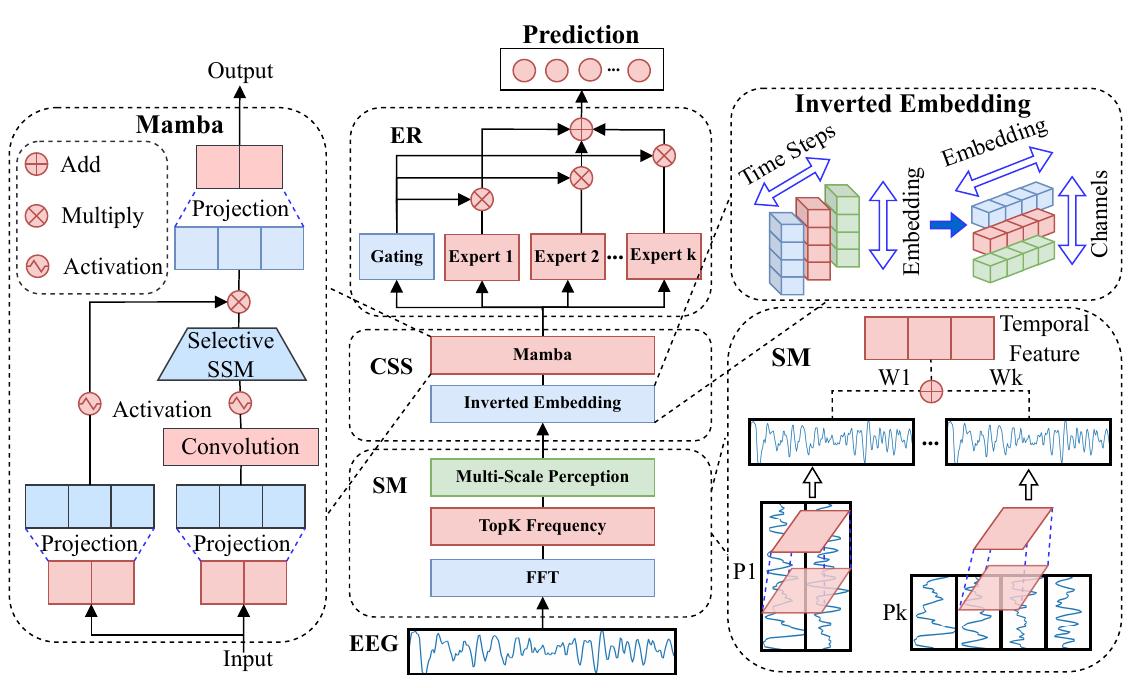}
	\caption{Overview of the prioritized channel-importance module. The center column, from bottom to top, comprises Seasonality Mining (SM), Channelwise State Space (CSS), and Expert Router (ER). The left panel shows the schematic of the Mamba block. The right panel, from bottom to top, shows the multi-scale feature fusion module and the inverted-embedding module. (SSM: State Space Model, FFT: Fast Fourier Transform.)}
	\label{fig.mamba}
\end{figure*}

In this section, we will introduce PRISM, which is composed of two modules: 
(i) a prioritized channel-importance module, and 
(ii) a semi-supervised domain-adaptation module. 
As illustrated in Fig.~\ref{fig.mamba}, the prioritized channel-importance module is implemented in three stages, namely Seasonality Mining (SM), Channelwise State Space (CSS), and Expert Router (ER). 
Fig.~\ref{fig.prism} depicts the semi-supervised domain-adaptation module tailored for cross-subject EEG emotion recognition, which integrates weak and strong augmentations, confidence-thresholded pseudo labeling, consistency regularization, entropy minimization, and a feature distribution alignment term for domain adaptation.

\subsection{Prioritized Channel Importance}

\subsubsection{Seasonality Mining}

Seasonal or scale-specific temporal cues are informative for sequence modeling \cite{wu2022timesnet, zhou2022fedformer}. 
As shown in Fig.~\ref{fig.mamba}, we extract multi-scale temporal representations from an EEG segment \(x\in\mathbb{R}^{L\times D}\) (length \(L\), channels \(D\)) in three steps: frequency-guided scale selection, blockwise multi-scale perception, and weighted fusion.

\paragraph{Frequency-guided scale selection.}
Let \(\mathcal{F}\) denote the fast Fourier transform and \(\mathcal{A}\) the amplitude operator.
We compute the spectrum \(A=\mathcal{A}(\mathcal{F}(x))\), select the top-\(K\) prominent frequencies \(\{f_i\}_{i=1}^{K}=\mathrm{TopK}(A)\), and convert them to periods $p_i = \Big\lfloor \frac{L}{f_i}\Big\rfloor.$
Nonnegative scale weights are obtained by a softmax over spectral amplitudes:
\begin{equation}
	w_i \;=\; \frac{\exp(\mathcal{A}(f_i))}{\sum_{j=1}^{K}\exp(\mathcal{A}(f_j))}, 
	\quad i=1,\dots,K .
	\label{eq:seasonality-weights}
\end{equation}

\paragraph{Blockwise multi-scale perception (MSP)}
For each period \(p_i\), we pad \(x\) to a length divisible by \(p_i\), then reshape the sequence into 2-D blocks through a period-wise rearrangement operator \(\mathcal{R}_{p_i}\):
\begin{equation}
	X^{(i)}_{2\mathrm{D}} \;=\; \mathcal{R}_{p_i}\!\big(\mathrm{Pad}_{p_i}(x)\big)
	\;\in\; \mathbb{R}^{p_i \times q_i \times D},
\end{equation}
where \(q_i\) is the number of blocks after padding. 
(\textit{Note: the subscripts ``1D/2D'' indicate the number of temporal axes only, and the channel axis \(D\) is always present in the tensor shape but omitted in the subscript for brevity.})
On these blocks, we apply a multi-scale perception (MSP) operator with kernel set \(\{K_m\}_{m=1}^{M}\),
\begin{equation}
	\widetilde{X}^{(i)}_{2\mathrm{D}} \;=\; \sum_{m=1}^{M} \mathrm{Conv}_{K_m}\!\big(X^{(i)}_{2\mathrm{D}}\big),
\end{equation}
and fold the result back to a one-dimensional time–channel layout:
\begin{equation}
	x^{(i)}_{1\mathrm{D}} \;=\; 
	\mathcal{R}^{-1}_{p_i}\!\big(\widetilde{X}^{(i)}_{2\mathrm{D}}\big)
	\;\in\; \mathbb{R}^{L\times D}.
\end{equation}

\paragraph{Multi-scale fusion.}
Finally, we fuse the per-scale representations using the weights in \eqref{eq:seasonality-weights}:
\begin{equation}
	x_{\mathrm{ms}} \;=\; \sum_{i=1}^{K} w_i\, x^{(i)}_{1\mathrm{D}}
	\;\in\; \mathbb{R}^{L\times D}.
	\label{eq:xms}
\end{equation}
The tensor \(x_{\mathrm{ms}}\) serves as the input to the subsequent Channelwise State Space stage.

\subsubsection{Channelwise State Space}
\label{subsec:css}

EEG channels recorded at the same time step may correspond to different neural events. Some channels can be at a peak while others are at a trough. Mapping signals from different channels at the same time into a single token risks mixing heterogeneous events \cite{zhou2025mimamba}. Moreover, a single time step rarely captures a complete event \cite{liu2023itransformer}. Motivated by these considerations, we adopt an inverted embedding scheme: instead of forming tokens by concatenating channels at the same time step (the conventional choice), we form tokens by concatenating the temporal trajectory of a single channel. This preserves channel structure and strengthens long-range temporal modeling. Formally, let \(x_{\mathrm{ms}}\in\mathbb{R}^{L\times D}\) be the output of Seasonality Mining. 
We exchange the time and channel axes using a permutation operator \(\mathrm{SwapAxes}_{L, D}\) (it swaps axis \(L\) with axis \(D\)):
\begin{equation}
	\widehat{x} \;=\; \mathrm{SwapAxes}_{L,D}\!\big(x_{\mathrm{ms}}\big) \;\in\; \mathbb{R}^{D\times L}.
\end{equation}

A Mamba \cite{gu2023mamba} block \(m_{\theta}(\cdot)\) is then applied in this channel-token space to capture spatiotemporal interactions, and the result is mapped back to the time–channel layout:
\begin{equation}
	\tilde{h} \;=\; \mathrm{SwapAxes}_{D,L}\!\big(m_{\theta}(\widehat{x})\big) \;\in\; \mathbb{R}^{L\times D}.
\end{equation}
We treat \(m_{\theta}\) as an encoder here. Mamba views a one dimensional sequence as a process driven by a continuous time dynamical system. Compared with the quadratic complexity of attention, Mamba performs training and inference with nearly linear complexity, which makes it suitable for EEG signals that span multiple temporal scales. Concretely, an input \(x(t)\in\mathbb{R}\) evolves through a hidden state \(h(t)\in\mathbb{R}^{d}\) and produces an output \(y(t)\in\mathbb{R}\). The evolution is controlled by three parameter matrices \(A\in\mathbb{R}^{d\times d}\), \(B\in\mathbb{R}^{d\times 1}\), and \(C\in\mathbb{R}^{1\times d}\), namely
\begin{equation}
	\frac{d}{dt}h(t)=A\,h(t)+B\,x(t),
	\quad
	y(t)=C\,h(t).
\end{equation}

Real world time series are discrete. Mamba therefore adopts a zero-order hold discretization via time scale parameter $\Delta$ and obtains the discrete parameters and the new recursion:
\begin{equation}
	\overline{A}=\exp(\Delta A),\quad
	\overline{B}=(\Delta A)^{-1}\big(\exp(\Delta A)-I\big)\,\Delta B,
\end{equation}
\begin{equation}
	h_{t}=\overline{A}\,h_{t-1}+\overline{B}\,x_{t},
	\quad
	y_{t}=C\,h_{t}.
\end{equation}

To enable parallelization, the entire mapping can be written as a single structured convolution. The convolution kernel and the output are:
\begin{equation}
	\widehat{K}
	=
	\big(
	C\,\overline{B},
	\ C\,\overline{A}\,\overline{B},
	\ \ldots,
	\ C\,\overline{A}^{\,L-1}\overline{B}
	\big),
	\quad
	y=x * \widehat{K},
\end{equation}
where \(\widehat{K}\) is a structured convolution kernel generated from \(A\), \(B\), and \(C\). This formulation enables efficient parallel convolution for long sequences while preserving the capacity to capture long range temporal dependencies.

\subsubsection{Expert Router}
After Seasonality Mining and Channelwise State Space, we obtain an EEG representation
that captures long-range temporal dependencies and fine-grained spatiotemporal
interactions. We then introduce an expert router to prioritize channel importance.
As shown in the dashed box (middle-top) of Fig.~\ref{fig.mamba}, the $i$-th expert consists
of a channel-weight vector $c_i\!\in\!\mathbb{R}^{D}$ and a channel mapping network
$\phi_i:\mathbb{R}^{D}\!\to\!\mathbb{R}^{D}$ implemented by a two-layer MLP. For any
time index $t$,
\begin{equation}
	\label{eq:8}
	u_i(t) = \tilde{h}(t)\odot c_i, \qquad
	E_i(t) = \phi_i\!\big(u_i(t)\big).
\end{equation}
Stacking over time yields $E_i(\tilde{h})\in\mathbb{R}^{L\times D}$, where each expert
learns a specific channel-weighting composition. In parallel, we summarize the temporal
dimension by a mean operator to obtain a time-averaged descriptor
$\mu=\tfrac{1}{L}\sum_{t=1}^{L}\tilde{h}(t)\in\mathbb{R}^{D}$ and compute noise-free
expert logits $\ell=W_{\text{gate}}\mu\in\mathbb{R}^{E}$. During training, Gaussian noise is injected to stabilize routing and to prevent the model
from collapsing onto a single expert:
\begin{equation}
	\sigma=\mathrm{softplus}\!\big(W_{\text{noise}}\mu\big)+\varepsilon_{0},
	\tilde{\ell}=\ell+\epsilon,
	\epsilon\sim\mathcal{N}\!\big(0,\mathrm{diag}(\sigma^{2})\big),
\end{equation}


where $\varepsilon_{0}$ is a constant and $\epsilon$ is Gaussian noise. At inference time we use the noise-free logits $\ell$ for stable predictions. We then
select the top-$k$ experts $S=\mathrm{TopK}(\tilde{\ell},k)$ and normalize on the selected
indices:
\begin{equation}
	s_{S}=\mathrm{softmax}(\tilde{\ell}_{S}), \quad
	s_{j}=0 \;\; (j\notin S).
\end{equation}

The final routed representation is a weighted mixture of expert outputs:
\begin{equation}
	y \;=\; \sum_{i=1}^{E} s_i\, E_i(\tilde{h}) \;\in\; \mathbb{R}^{L\times D}.
\end{equation}
A downstream classification head takes $y$ to produce predictions. Using multiple experts
enables a diverse set of channel-weight combinations rather than a single fixed pattern.
$\{c_i\}$ realize channel-wise soft prioritization, while $s$ provides sample-adaptive
expert mixing. The router is fully data-driven, and the expert parameters and channel
weights are learned end-to-end jointly with the rest of the network.

\begin{figure*}[t]
	\centering
	\includegraphics[width=0.7\linewidth]{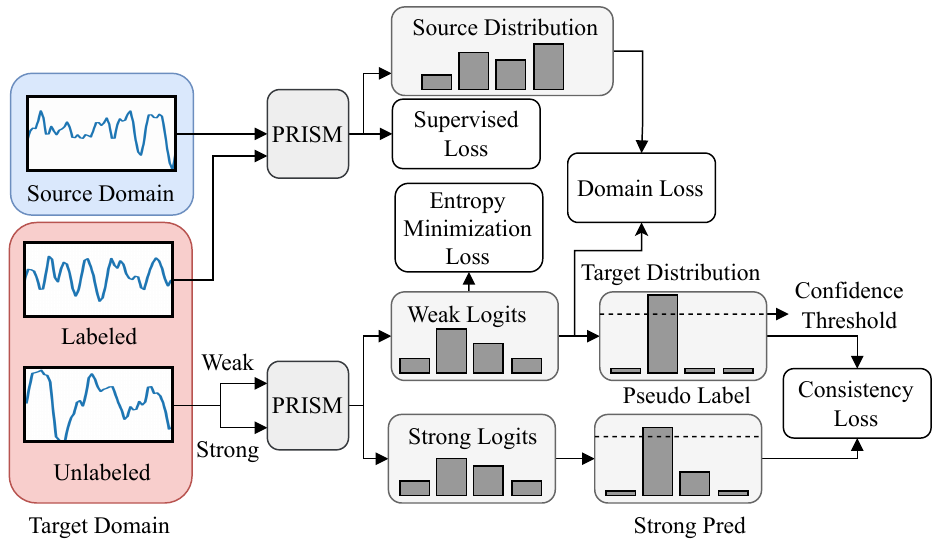}
	\caption{Pipeline of semi-supervised domain adaptation for EEG. Blue and red blocks denote source-domain and target-domain data, respectively. PRISM indicates our classifier (pluggable and replaceable). The learning objective includes four terms: supervised loss, entropy minimization, consistency regularization, and domain alignment.}
	\label{fig.prism}
\end{figure*}

\subsection{Semi-supervised domain adaptation for EEG}
In this subsection, we propose the semi-supervised domain adaptation used for EEG emotion recognition. The overall pipeline is shown in Fig. \ref{fig.prism}. 
To enhance the effective capacity of the model while remaining label-efficient, we generate two views of each target sample with weak and strong augmentations that are tailored to EEG. Let $a_w$ and $a_s$ denote the weak and strong augmentations, respectively. 
They are defined as:
\begin{align}
	a_w(x) &= x + \epsilon,  \epsilon \sim \mathcal{N}(0, \sigma_w^2), \\
	a_s(x) &= x + \epsilon' + \delta_{\mathrm{drop}} + \delta_{\mathrm{jitter}} .
\end{align}
$\epsilon$ and $\epsilon'$ are both Gaussian noise. $\delta_{\mathrm{drop}}$ is a channel-wise random zero mask and $\delta_{\mathrm{jitter}}$ is a perturbation along the temporal axis. 
The weak view preserves the main structure of the original signal, whereas the strong view combines multiple perturbations to improve robustness. For labeled samples $(x^\ell, y^\ell)$, we minimize $\mathcal{L}_{\mathrm{sup}} = \mathrm{CE}\!\big(z(x^\ell),\, y^\ell\big)$, where $z(\cdot)$ denotes the network logits. For an unlabeled target sample $x^u$, we compute the weak-view logits and probabilities as follows:
\begin{equation}
	z_w = z\!\big(a_w(x^u)\big), 
	\quad 
	p_w = \mathrm{softmax}(z_w).
\end{equation}
Obtaining the hard pseudo label $\hat{y}=\arg\max p_w$, and build a confidence mask $m = \mathbf{1}\{\max p_w \ge \tau\}$. Only high-confidence samples contribute to the consistency objective. 
With the strong-view logits $z_s = z\!\big(a_s(x^u)\big)$, the loss is:
\begin{equation}
	\mathcal{L}_{\mathrm{cons}} 
	= \frac{1}{\|m\|_{1}} \sum m \cdot \mathrm{CE}\!\big(z_s,\, \hat{y}\big).
\end{equation}

To encourage confident predictions on the weak view, we minimize:
\begin{equation}
	\mathcal{L}_{\mathrm{ent}}
	= \frac{1}{C}\sum_{c=1}^{C} \big[-\, p_w^{(c)} \log p_w^{(c)} \big].
\end{equation}

Due to the source and target batches not being identically distributed in the cross-subject setting, which degrades generalization \cite{zhou2024eegmatch}. 
We align the mean predictive distributions of the two domains:
\begin{equation}
	\bar{p}_s = \mathrm{mean}\!\big(\mathrm{softmax}(z(x^s))\big), 
	\bar{p}_t = \mathrm{mean}\!\big(\mathrm{softmax}(z(a_w(x^u)))\big),
\end{equation}
\begin{equation}
	\mathcal{L}_{\mathrm{dom}} 
	= \big\| \bar{p}_s - \bar{p}_t \big\|_2^{2}.
\end{equation}
The mean operator is taken over the minibatch. Finally, the total loss combines all terms with nonnegative weights $\lambda_{\mathrm{cons}}$, $\lambda_{\mathrm{ent}}$ and $\lambda_{\mathrm{dom}}$ as follows:
\begin{equation}
	\mathcal{L} 
	= \mathcal{L}_{\mathrm{sup}}
	+ \lambda_{\mathrm{cons}} \mathcal{L}_{\mathrm{cons}}
	+ \lambda_{\mathrm{ent}} \mathcal{L}_{\mathrm{ent}}
	+ \lambda_{\mathrm{dom}} \mathcal{L}_{\mathrm{dom}} .
\end{equation}

\section{Experiment and Results Analysis}\label{sec4}

\subsection{Datasets and preprocessing}
We systematically evaluate PRISM on three public EEG emotion datasets, DEAP~\cite{koelstra2011deap}, DREAMER~\cite{katsigiannis2017dreamer}, and SEED~\cite{zheng2015investigating}. 
DEAP contains 32 participants who watched music videos to induce emotions, and 32-channel EEG was recorded for each participant. 
DREAMER provides 14-channel EEG from 23 participants. 
Both DEAP and DREAMER include emotion annotations along the valence and arousal dimensions. SEED contains recordings from 15 participants collected in three sessions with 62 channels, and it provides three discrete emotion categories (positive, neutral, and negative). For preprocessing, DEAP and DREAMER were downsampled to 128\,Hz and filtered with a 4–45\,Hz bandpass. 
SEED was downsampled to 200\,Hz and filtered to 0–75\,Hz. 
All datasets were segmented into 1\,s nonoverlapping windows, and z-score standardization was applied per channel.

\subsection{Baselines and evaluation}
We compare against six advanced time series models that are widely used as baselines from Time-Series-Library \footnote{Baseline implementations are taken from the public repository at \url{https://github.com/thuml/Time-Series-Library}.}: iTransformer~\cite{liu2023itransformer}, DLinear~\cite{zeng2023transformers}, TimesNet~\cite{wu2022timesnet}, NTransformer~\cite{liu2022non}, Informer~\cite{zhou2021informer}, and TCN~\cite{bai2018empirical}. To assess performance and generalization, we adopt three protocols, inter-subject, cross-subject and subject-dependent. In the inter-subject setting, we pool data from all subjects, shuffle, and split it into training and test sets with a $3{:}1$ ratio. In the cross-subject setting for semi-supervised adaptation, we construct a disjoint target-domain subset comprising $30\%$, $20\%$, and $10\%$ of participants on DEAP, DREAMER, and SEED, respectively. For each target subject, only 30\% samples are annotated. For subject-dependent case, model outputs prediction individually and 25\% of samples for each subject are used for testing. SEED contains three sessions per participant. We therefore report inter-session results and also evaluate each session independently. Classification accuracy is used as the primary metric since our sample distribution for each class is balanced.

\subsection{Implementation details}
\label{app:implementation}
We implement PRISM in PyTorch and run all experiments on two NVIDIA RTX 4090 GPUs.
For DEAP and DREAMER, the length of a single sample is \(128\), and for SEED, it is \(200\). We use Adam optimizer with an initial learning rate of \(1\times 10^{-4}\), a batch size of \(32\), and train for \(10\) epochs. In seasonality mining, we retain the two scales with the highest spectral amplitudes \((K=2)\). The expert router instantiates up to eight experts and selects the top four for each sample \((E=8,\ k=4)\). Loss weights are set to \(\lambda_{\mathrm{cons}}=1\), \(\lambda_{\mathrm{ent}}=0.1\), and \(\lambda_{\mathrm{dom}}=0.1\). The confidence threshold is \(\tau=0.95\).

\begin{table}[t]
	\centering
	\caption{Inter-subject accuracy (\%). The best results are in bold and the second-best are underlined.}
	\label{tab:inter_subject}
	\renewcommand{\arraystretch}{0.95}
	\setlength{\tabcolsep}{7pt}
	\begin{tabularx}{\linewidth}{P{0.16\linewidth} @{\hspace{8pt}} *{8}{Y}}
		\toprule
		\multirow{2}{*}{\centering Method} &
		\multicolumn{2}{c}{DEAP} &
		\multicolumn{2}{c}{DREAMER} &
		\multicolumn{4}{c}{SEED} \\
		\cmidrule(lr){2-3}\cmidrule(lr){4-5}\cmidrule(lr){6-9}
		& V & A & V & A & Inter & S0 & S1 & S2 \\
		\midrule
		iTransformer & 76.63 & 78.19 & 77.50 & 82.73 & 57.47 & 81.57 & 78.76 & 71.93 \\
		DLinear      & 80.61 & 82.47 & \secondbest{81.57} & \secondbest{86.03} & 41.78 & 48.87 & 45.90 & 43.88 \\
		TimesNet     & 85.75 & \secondbest{87.96} & 80.25 & 85.28 & 70.50 & 86.90 & 86.51 & 80.95 \\
		NTransformer & 82.61 & 85.01 & 78.56 & 83.86 & 60.75 & 81.08 & 81.12 & 73.76 \\
		Informer     & 81.79 & 83.56 & 80.36 & 84.25 & 51.96 & 66.53 & 65.97 & 58.63 \\
		TCN          & \secondbest{86.56} & 87.78 & 78.90 & 85.16 & \secondbest{74.65} & \secondbest{92.03} & \secondbest{92.39} & \secondbest{85.25} \\
		Ours         & \best{90.35} & \best{91.65} & \best{90.14} & \best{92.53} & \best{97.62} & \best{97.50} & \best{97.59} & \best{97.07} \\
		\bottomrule
	\end{tabularx}
	    \begin{justify}
		\footnotesize
		\textit{Note:} V: valence; A: arousal; Inter: inter-session; S0, S1, S2 represents session 0, 1, 2, respectively. The other tables are the same.
	\end{justify}
\end{table}

\subsection{Experimental results}
\subsubsection{Inter-subject results}
As shown in Table~\ref{tab:inter_subject}, PRISM achieves the best and most stable performance across all datasets and settings. On DEAP, PRISM surpasses TCN by 3.79\% on valence and TimesNet by 3.69\% on arousal.
On DREAMER, the margins over the second best are 8.57\% for valence and 6.50\% for arousal.
The gains are largest on SEED. Under the inter setting PRISM exceeds TCN by 22.97\%, and across sessions S0, S1, and S2 the margins are 5.47\%, 5.20\%, and 11.82\%, respectively.
Compared with DEAP and DREAMER, SEED has more channels and multiple recording sessions, which yields stronger channel redundancy and cross-session variation.
PRISM benefits most in this regime because it highlights stable electrodes while suppressing noisy or redundant ones.
Although TCN is stronger than most baselines on SEED, it still struggles with the large channel count and session variability.
DLinear is relatively strong on DREAMER, indicating that trend and seasonal components can fit a reasonable decision boundary.
PRISM nevertheless improves on this baseline through multi-expert channel weighting and multi-scale temporal modeling, providing additional discriminative power.

\begin{table}[t]
	\centering
	\caption{Cross-subject accuracy (\%). The best results are in bold and the second-best are underlined.}
	\label{tab:cross_subject}
	\renewcommand{\arraystretch}{0.95}
	\setlength{\tabcolsep}{7pt}
	\begin{tabularx}{\linewidth}{P{0.16\linewidth} @{\hspace{8pt}} *{8}{Y}}
		\toprule
		\multirow{2}{*}{\centering Method} &
		\multicolumn{2}{c}{DEAP} &
		\multicolumn{2}{c}{DREAMER} &
		\multicolumn{4}{c}{SEED} \\
		\cmidrule(lr){2-3}\cmidrule(lr){4-5}\cmidrule(lr){6-9}
		& V & A & V & A & Inter & S0 & S1 & S2 \\
		\midrule
		iTransformer & 66.48 & 65.97 & 65.68 & 83.33 & 42.40 & 56.52 & 57.71 & 58.31 \\
		DLinear      & 73.63 & 72.60 & \secondbest{69.79} & 83.22 & 36.25 & 40.37 & 38.64 & 39.78 \\
		TimesNet     & \secondbest{77.26} & 77.76 & 69.77 & \secondbest{86.77} & 48.33 & 54.39 & 60.08 & 57.33 \\
		NTransformer & 71.89 & 69.08 & 69.25 & 84.26 & 45.06 & 56.43 & 52.94 & 52.73 \\
		Informer     & 70.80 & 69.31 & 68.45 & 84.97 & 42.73 & 50.49 & 48.03 & 43.23 \\
		TCN          & 77.06 & \secondbest{79.12} & 69.08 & 84.88 & \secondbest{54.58} & \secondbest{72.90} & \secondbest{69.80} & \secondbest{65.90} \\
		Ours         & \best{87.28} & \best{87.45} & \best{84.69} & \best{92.62} & \best{93.17} & \best{93.64} & \best{94.40} & \best{94.87} \\
		\bottomrule
	\end{tabularx}
\end{table}

\subsubsection{Cross-subject results}
Table~\ref{tab:cross_subject} reports the cross-subject results. 
Since individual variability and domain shift, all baselines drop notably compared with the inter-subject setting, whereas PRISM remains clearly ahead on every dataset and evaluation dimension. 
On DEAP, PRISM reaches 87.28\% on valence and 87.45\% on arousal, exceeding the second best by 10.02\% and 8.33\%, respectively. 
On DREAMER, PRISM attains 84.69\% on valence and 92.62\% on arousal, which are higher than DLinear at 69.79\% and TimesNet at 86.77\% by 14.90\% and 5.85\%. 
On SEED, the margins over the second best exceed 20\% in all sessions (Inter, S0, S1 and S2). 
We attribute the consistent advantage to three complementary factors. 
\textbf{First}, channel prioritization suppresses weak or noisy electrodes and highlights stable, emotion-relevant spatial signals. 
\textbf{Second}, inverted embedding combined with a state-space backbone captures longer-range, multi-scale spatiotemporal structure, which stabilizes representations under large across-subject variation. 
\textbf{Third}, the semi-supervised adaptation module uses a confidence threshold of 95\% for pseudo labels, entropy minimization on weak views, and source–target alignment, thereby reducing pseudo-label noise and mitigating domain shift. 
Although TCN remains the strongest baseline on many settings, indicating the value of local temporal inductive bias, PRISM consistently surpasses it, especially on DEAP valence and across all SEED protocols. 
TimesNet leads among baselines on DREAMER arousal, suggesting stronger periodic or multi-scale components in this dimension, yet PRISM still achieves the best overall results.

\begin{table}[t]
	\centering
	\caption{Subject-dependent accuracy (\%). Best is shown in bold.}
	\label{tab:intra_subject}
	\renewcommand{\arraystretch}{0.6}
	\setlength{\tabcolsep}{7pt}
	\begin{tabularx}{\linewidth}{P{0.16\linewidth} @{\hspace{8pt}} *{8}{Y}}
		\toprule
		\multirow{2}{*}{\centering Method} &
		\multicolumn{2}{c}{DEAP} &
		\multicolumn{2}{c}{DREAMER} &
		\multicolumn{4}{c}{SEED} \\
		\cmidrule(lr){2-3}\cmidrule(lr){4-5}\cmidrule(lr){6-9}
		& V & A & V & A & Inter & S0 & S1 & S2 \\
		\midrule
		iTransformer   & 94.59 & 94.59 & 93.58 & 95.09 & 88.19 & 93.05 & 94.01 & 89.00 \\
		DLinear        & 96.43 & 96.78 & \best{97.94} & 97.94 & 69.65 & 94.00 & 95.32 & 93.91 \\
		TimesNet       & 97.36 & 97.65 & 97.84 & 97.88 & 93.36 & 96.77 & 96.31 & 93.07 \\
		NTransformer   & 97.38 & 97.45 & 96.98 & 97.62 & 90.07 & 95.95 & 95.87 & 91.04 \\
		Informer       & \best{97.61} & \best{97.94} & 97.83 & \best{98.15} & 86.55 & 95.62 & 95.65 & 92.27 \\
		TCN            & 95.36 & 95.52 & 93.56 & 95.23 & 92.43 & 94.11 & 94.70 & 87.05 \\
		Ours           & 96.23 & 96.64 & 96.94 & 96.91 & \best{96.52} & \best{97.28} & \best{97.10} & \best{94.99} \\
		\bottomrule
	\end{tabularx}
\end{table}

\subsubsection{Subject-dependent results}
We also perform experiments under the subject-dependent setting, where a separate model is trained and evaluated for each participant. The results are shown in Table~\ref{tab:intra_subject}. \textbf{First}, on DEAP and DREAMER the overall performance is already near a ceiling, with most methods in the range of 96\% - 98\%. The performance gaps are therefore compressed, which suggests that within a single subject the emotion related temporal patterns are relatively consistent and the task behaves like a standard sequence classification problem, where complex cross domain alignment is not the key factor. Models that rely on attention or multi-scale convolutions, such as Informer and DLinear, tend to be slightly ahead, while PRISM is comparable but not dominant on these datasets, which is consistent with the fact that PRISM is not designed specifically for the subject-dependent scenarios. \textbf{Second}, PRISM shows the clearest advantage on SEED. Whether we use the inter split or the three independent sessions, PRISM achieves the highest accuracy. This aligns with the characteristics of SEED, which has many channels and larger variation across sessions. The results indicate that even within a subject, PRISM brings stable gains across sessions by reducing redundancy and noise. \textbf{Finally}, DLinear remains strong on DREAMER, which implies that trend and seasonal components can model within subject emotion signals well. Overall, the subject-dependent setting emphasizes the precise modeling of a single subject’s stable patterns, while PRISM provides the most value when channel dimensionality is high and session variability is large.

\begin{table}[t]
	\centering
	\caption{Average classification accuracy (\%) compared with the state-of-the-art methods on three datasets}
	\begin{tabular}{p{1.6cm}p{1.2cm}p{1.2cm}p{0.7cm}p{1cm}p{0.7cm}}
		\toprule
		Model& Technique & Mode & DEAP& DREAMER & SEED \\
		\midrule
		TAE\cite{cheng2024novel} & Static & Dependent & 66.29 & - & - \\
		miMamba\cite{zhou2025mimamba} & Static  & Dependent & 85.99 &84.47  & 86.10\\
		CSGNN\cite{lin2023eeg} & Dynamic  & Independent & 75.88 & - & 81.85\\
		ARCNN\cite{tao2020eeg} & Dynamic  & Dependent & 93.55  & \textbf{97.96} & 83.93\\
		CWGCN\cite{yang2024automatically} & Dynamic  & Dependent & - & - & 94.97\\
		\textbf{PRISM} & Dynamic  & Dependent & \textbf{96.44} & 96.93 & \textbf{96.52}\\
		\midrule
		DSSN\cite{li2024dynamic} & Dynamic Stream & Independent & 59.76 & 63.40 & - \\
		TSF\cite{wang2024cascaded} & Self-supervised  & Independent & 67.59 &70.34  & -\\
		GUSA\cite{li2024gusa} & Unsupervised  & Independent & - & - & 91.77\\
		GDDN\cite{chen2024gddn} & Graph  & Independent & - & - & 92.54\\
		EmT\cite{ding2025emt} & Transformer  & Independent & - & - & 80.20\\
		EEGMatch\cite{zhou2024eegmatch} & Semi-supervised  & Independent & - & - & 91.35\\
		DS-AGC\cite{ye2024semi} & Semi-supervised & Independent & - & - & 87.37 \\
		TAS-Net\cite{zhang2023unsupervised} & Unsupervised  & Independent & 59.18 &-  & 63.10\\
		CGRU-MDGN\cite{guo2024convolutional} & Graph  & Independent & 70.00 & 85.39 & 90.40\\
		LSTM-CNN\cite{rajpoot2022subject} & Attention  & Independent & 67.70 & - & 76.70\\
		\textbf{PRISM} & Semi-supervised  & Independent & \textbf{86.04} & \textbf{88.85} & \textbf{93.17}\\
		\bottomrule
	\end{tabular}
	\begin{justify}
		\footnotesize
		\textit{Note:} 'Static' means manually pre-selected channels; 'Dynamic' means model-driven, learned channel weighting. 'Mode' denotes the evaluation protocol: Dependent = subject-dependent setting; Independent = cross-subject setting.
	\end{justify}
	\label{tab:sota}
\end{table}

\subsubsection{Comparison with state-of-the-art methods}
Table \ref{tab:sota} presents a comparative analysis of PRISM against state-of-the-art methods, categorized into channel selection and domain adaptation (DA) techniques. The top section compares methods based on channel selection, distinguishing between static selection (manually pre-determined channels) and dynamic selection (weights automatically assigned by model). The bottom section benchmarks various domain adaptation paradigms, including semi-supervised, unsupervised, graph-based and other approaches. The results clearly indicate that methods employing dynamic channel selection generally outperform static selection approaches, underscoring the necessity of data-dependent channel prioritization to mitigate redundancy. Crucially, PRISM achieves the best overall performance on the DEAP and SEED datasets. Although it is slightly outperformed by ARCNN on DREAMER, the margin is negligible (only 1.03 points). In the challenging subject-independent setting, PRISM’s combination of prioritized channel importance and semi-supervised domain adaptation consistently yields both superior accuracy compared to unsupervised or graph-based DA methods. Furthermore, even in the subject-dependent setting, PRISM maintains performance that is either comparable to or better than the existing optimal methods. These results highlight the robust and synergistic efficacy of PRISM’s two core modules in addressing the fundamental challenges of inter-subject variability and channel redundancy in EEG emotion recognition.

\subsection{Ablation studies}
\subsubsection{Inter-subject}
\begin{table}[t]
	\centering
	\caption{Ablation studies on inter-subject accuracy (\%). Best is shown in bold. ER: Expert Router, CSS: Channelwise State Space, SM: Seasonality Mining.}
	\label{tab:ablation_inter}
	\renewcommand{\arraystretch}{0.6}
	\setlength{\tabcolsep}{7pt}
	\begin{tabularx}{\linewidth}{P{0.18\linewidth} @{\hspace{8pt}} *{8}{Y}}
		\toprule
		\multirow{2}{*}{\centering Variant} &
		\multicolumn{2}{c}{DEAP} &
		\multicolumn{2}{c}{DREAMER} &
		\multicolumn{4}{c}{SEED} \\
		\cmidrule(lr){2-3}\cmidrule(lr){4-5}\cmidrule(lr){6-9}
		& V & A & V & A & Inter & S0 & S1 & S2 \\
		\midrule
		w/o ER       & 86.49 & 87.23 & 87.02 & 89.42 & 91.96 & 95.43 & 95.09 & 91.08 \\
		w/o CSS      & 87.77 & 89.50 & 81.67 & 87.79 & 63.89 & 93.83 & 94.24 & 87.41 \\
		w/o SM       & \best{90.48} & 91.13 & 88.72 & 91.24 & 88.88 & 95.95 & 93.75 & 90.70 \\
		w/o CSS+SM   & 84.23 & 86.15 & 78.33 & 83.54 & 66.88 & 83.58 & 78.07 & 71.14 \\
		w/o ER+CSS   & 69.94 & 71.41 & 71.60 & 77.71 & 54.21 & 71.35 & 65.13 & 56.02 \\
		w/o ER+SM    & 86.08 & 86.60 & 85.33 & 88.47 & 81.90 & 90.56 & 89.38 & 78.49 \\
		\midrule
		Full         & 90.35 & \best{91.65} & \best{90.14} & \best{92.53} & \best{97.62} & \best{97.50} & \best{97.59} & \best{97.07} \\
		\bottomrule
	\end{tabularx}
	
\end{table}

As shown in Table~\ref{tab:ablation_inter}, we report the impact of removing each module under the inter-subject setting. The three modules play distinct roles and also reinforce one another, and CSS is the most critical component. Removing CSS drops SEED-inter from 97.62\% to 63.89\%, and all three sessions also decline markedly. This indicates that without explicit modeling of spatiotemporal structure, redundancy and noise are amplified. ER delivers steady gains. When ER is removed, the PRISM will degenerate into miMamba \cite{zhou2025mimamba} using static channel selection. Without ER, SEED-inter remains at 91.96\% but is clearly lower than the full model, and S1 and S2 decrease to 95.09\% and 91.08\%. This shows that soft routing is an effective unified mechanism across datasets for suppressing channel redundancy. SM improves generalization overall, especially on SEED where the number of channels is large and cross-session variation is strong. Removing SM reduces SEED inter from 97.62\% to 88.88\%. There is a small reversal on DEAP valence, where 90.48\% slightly exceeds 90.35\%. This likely occurs when samples are short, channels are fewer, or the periodic structure is weak, in which case explicit multi-scale seasonal modeling brings limited benefit and may overlap with other submodules. More importantly, removing two modules at the same time leads to structural collapse. Removing CSS and SM yields 66.88\% on SEED inter, which suggests that the model is left without multi-scale temporal cues and without channel-state constraints, and thus relies almost only on a lightweight expert ensemble and cannot resist cross-subject shift. The degradation is most severe when ER and CSS are both removed, which confirms that the combination of soft channel selection and channelwise temporal modeling is the core defense against channel redundancy.

\subsubsection{Cross-subject}
\begin{table}[t]
	\centering
	\caption{Ablation studies on cross-subject accuracy (\%). Best is shown in bold. ER: Expert Router, CSS: Channelwise State Space, SM: Seasonality Mining.}
	\label{tab:ablation_cross}
	\renewcommand{\arraystretch}{0.6}
	\setlength{\tabcolsep}{7pt}
	\begin{tabularx}{\linewidth}{P{0.18\linewidth} @{\hspace{8pt}} *{8}{Y}}
		\toprule
		\multirow{2}{*}{\centering Variant} &
		\multicolumn{2}{c}{DEAP} &
		\multicolumn{2}{c}{DREAMER} &
		\multicolumn{4}{c}{SEED} \\
		\cmidrule(lr){2-3}\cmidrule(lr){4-5}\cmidrule(lr){6-9}
		& V & A & V & A & Inter & S0 & S1 & S2 \\
		\midrule
		w/o ER       & 87.12 & 87.15 & 82.89 & 91.07 & 85.87 & 92.03 & 90.84 & 88.71 \\
		w/o CSS      & 85.35 & 85.80 & 71.39 & 86.42 & 53.58 & 77.93 & 79.64 & 78.16 \\
		w/o SM       & \textbf{90.18} & \textbf{90.30} & 83.66 & 91.57 & 82.02 & 85.17 & 88.39 & 79.66 \\
		w/o CSS+SM   & 78.08 & 79.44 & 67.19 & 84.56 & 49.72 & 62.83 & 58.92 & 55.67 \\
		w/o ER+CSS   & 65.01 & 62.71 & 61.94 & 79.71 & 41.32 & 51.39 & 45.30 & 44.35 \\
		w/o ER+SM    & 86.45 & 86.48 & 83.39 & 88.84 & 67.58 & 80.80 & 77.33 & 64.00 \\
		\midrule
		Full         & 87.28 & 87.45 & \textbf{84.69} & \textbf{92.62} & \textbf{93.17} & \textbf{93.64} & \textbf{94.40} & \textbf{94.87} \\
		\bottomrule
	\end{tabularx}
\end{table}

Table~\ref{tab:ablation_cross} reports the ablation studies under the cross-subject setting. The full model remains the top performance on DREAMER and SEED. Compared with the inter-subject results in Table \ref{tab:ablation_inter}, removing CSS causes a huge drop, indicating that CSS plays the key role in spatiotemporal feature extraction. Removing ER produces a consistent but moderate degradation, and the effect is more visible on SEED where the channel count and variability are higher. The effect of removing SM is data dependent. On DEAP it can match or slightly exceed the full model, whereas on DREAMER and SEED it generally degrades performance. Eliminating two modules leads to substantial deterioration, especially combinations that exclude CSS. This pattern mirrors Table \ref{tab:ablation_inter} but is amplified in the cross-subject regime, highlighting the complementarity of the three modules and the indispensable role of CSS. Overall, PRISM reaches its best performance through the synergy of the three modules, and weakening any two breaks this complementarity and causes a pronounced drop in accuracy.

\begin{table}[t]
	\centering
	\caption{Ablation studies on subject-dependent accuracy (\%). ER: Expert Router, CSS: Channelwise State Space, SM: Seasonality Mining. Best in each column is in bold.}
	\label{tab:ablation_intra}
	\renewcommand{\arraystretch}{0.6}
	\setlength{\tabcolsep}{7pt}
	\providecommand{\best}[1]{\textbf{#1}}
	\begin{tabularx}{\linewidth}{P{0.18\linewidth} @{\hspace{8pt}} *{8}{Y}}
		\toprule
		\multirow{2}{*}{\centering Variant} &
		\multicolumn{2}{c}{DEAP} &
		\multicolumn{2}{c}{DREAMER} &
		\multicolumn{4}{c}{SEED} \\
		\cmidrule(lr){2-3}\cmidrule(lr){4-5}\cmidrule(lr){6-9}
		& V & A & V & A & Inter & S0 & S1 & S2 \\
		\midrule
		w/o ER       & 95.14 & 95.19 & 95.09 & 96.04 & 93.51 & 94.62 & 93.07 & 87.71 \\
		w/o CSS      & 97.36 & \best{97.59} & \best{97.76} & \best{98.25} & 94.32 & 96.85 & 96.27 & 94.20 \\
		w/o SM       & 96.63 & 96.91 & 96.88 & 97.01 & 94.25 & 96.14 & 97.04 & 93.54 \\
		w/o CSS+SM   & \best{97.55} & 97.55 & 97.11 & 97.81 & 88.31 & 95.32 & 94.03 & 90.47 \\
		w/o ER+CSS   & 90.61 & 91.05 & 87.91 & 91.13 & 75.31 & 85.01 & 83.88 & 72.52 \\
		w/o ER+SM    & 94.89 & 95.24 & 95.06 & 95.99 & 87.08 & 91.72 & 90.56 & 82.89 \\
		Full         & 96.23 & 96.64 & 96.94 & 96.91 & \best{96.52} & \best{97.28} & \best{97.10} & \best{94.99} \\
		\bottomrule
	\end{tabularx}
\end{table}

\subsubsection{Subject-dependent}
Table~\ref{tab:ablation_intra} reports the ablation results under the subject-dependent setting. 
On SEED, the full PRISM consistently achieves the best performance, and removing any submodule leads to clear degradation. 
The drop is largest when both ER and CSS are removed, indicating that the combination of channelwise state modeling and soft routing is critical in regimes with many channels and strong cross-session variation. 
When removing CSS or ER alone, the changes on DEAP and DREAMER are small, whereas SEED still degrades, which suggests that explicit spatiotemporal modeling and channel routing are less beneficial for easier within-subject cases but become indispensable when channel redundancy and cross-session variability are stronger. 
SM mainly contributes to stability and refinement. Removing SM alone causes milder declines than removing CSS, but removing both SM and CSS produces a huge drop, showing that multi-scale temporal cues and channelwise state modeling are complementary. 
Overall, on datasets with many channels or large cross-session differences, the synergy among the three modules is irreplaceable.

\begin{table}[t]
	\centering
	\caption{Effect of top-$k$ channel filtering on cross-subject accuracy (\%). Better results are in bold.}
	\label{tab:topk_channels}
	\renewcommand{\arraystretch}{0.95}
	\setlength{\tabcolsep}{6pt}
	\begin{tabularx}{\linewidth}{P{0.16\linewidth} *{8}{Y}}
		\toprule
		\multirow{2}{*}{\centering Setting} &
		\multicolumn{2}{c}{DEAP} &
		\multicolumn{2}{c}{DREAMER} &
		\multicolumn{4}{c}{SEED} \\
		\cmidrule(lr){2-3}\cmidrule(lr){4-5}\cmidrule(lr){6-9}
		& V & A & V & A & Inter & S0 & S1 & S2 \\
		\midrule
		w/o top-$k$ & \textbf{87.36} & \textbf{87.46} & \textbf{85.53} & \textbf{93.38} & 87.96 & 92.00 & 93.37 & 93.97 \\
		with top-$k$    & 87.28 & 87.45 & 84.69 & 92.62 & \textbf{93.17} & \textbf{93.64} & \textbf{94.40} & \textbf{94.87} \\
		\bottomrule
	\end{tabularx}
\end{table}

\subsubsection{Effect of top-k channel filtering}

PRISM supports two implementations of the channel selection strategy. The first applies the weighting in Eq.~\ref{eq:8} to all channels and aggregates them by a weighted sum. The second selects the top-$k$ channels by the coefficients \(c_i\) in Eq.~\ref{eq:8} implementation (Results in Table \ref{tab:inter_subject}, \ref{tab:cross_subject}, \ref{tab:intra_subject}, \ref{tab:ablation_inter}, \ref{tab:ablation_cross} and \ref{tab:ablation_intra} are implemented by this way). To make the comparison explicit, Table~\ref{tab:topk_channels} reports results under a fixed \(k=4\) for the two settings with and without top-$k$ channel filtering. On DEAP and DREAMER the change after enabling top-$k$ is small and slightly negative. On SEED the gains are pronounced, most notably on the inter setting and consistently across the three sessions. These results indicate that top-$k$ channel filtering is more effective in regimes with many channels and stronger cross-session variation, where it suppresses redundant or session-specific noise and emphasizes stable electrodes. In datasets with fewer channels, hard filtering may discard weak yet useful signals. Consequently, the advantage of the channel selection strategy becomes more salient as the channel dimensionality increases.

\begin{table}[t]
	\centering
	\caption{Ablation on the number of experts and top-$k$ channels (accuracy \%). Best in each column is in bold.}
	\label{tab:experts_topk}
	\renewcommand{\arraystretch}{0.6}
	\setlength{\tabcolsep}{6pt}
	\begin{tabularx}{\linewidth}{P{0.07\linewidth} P{0.05\linewidth} *{5}{Y}}
		\toprule
		\multirow{2}{*}{\centering {Experts}} & \multirow{2}{*}{\centering Top-$k$} &
		\multicolumn{2}{c}{DEAP} & \multicolumn{2}{c}{DREAMER} & \multicolumn{1}{c}{SEED} \\
		\cmidrule(lr){3-4}\cmidrule(lr){5-6}\cmidrule(lr){7-7}
		& & V & A & V & A & Inter \\
		\midrule
		8  & 2  & 86.69 & 86.36 & 83.44 & 92.78 & 88.51 \\
		8  & 4  & \textbf{87.28} & \textbf{87.45} & 84.69 & 92.62 & 93.17 \\
		8  & 6  & 86.10 & 87.23 & 84.65 & 92.24 & 90.34 \\
		8  & 8  & 86.19 & 86.95 & 85.77 & \textbf{92.88} & 90.38 \\
		8  & 10 & 87.05 & 86.98 & 86.37 & 92.52 & 86.02 \\
		4  & 4  & 86.86 & 87.28 & \textbf{87.37} & 92.54 & 93.14 \\
		6  & 4  & 86.61 & 86.83 & 86.53 & 92.15 & \textbf{93.43} \\
		8  & 4  & \textbf{87.28} & \textbf{87.45} & 84.69 & 92.62 & 93.17 \\
		10 & 4  & 86.33 & 86.84 & 84.11 & 92.76 & 72.49 \\
		12 & 4  & 85.49 & 86.75 & 86.14 & 91.98 & 86.45 \\
		\bottomrule
	\end{tabularx}
\end{table}

\subsubsection{Sensitivity to the number of experts and top-$k$ channels.}

Table~\ref{tab:experts_topk} reports how the number of experts and the choice of top-$k$ channels affect accuracy. 
When the number of experts is fixed to 8, choosing a very small $k$ weakens the selectivity of the router, whereas a very large $k$ introduces redundancy. 
Aggregating results on DEAP, DREAMER, and SEED, $k=4$ is the most stable choice. 
The benefit is most evident on SEED, where the channel count is high and the across session variation is strong. 
When $k$ is fixed to 4, using too few experts limits the expressive power of routing, while too many experts can lead to unstable training. 
A smaller $k$ paired with a medium-sized expert set strikes a better balance between computation and accuracy. 
Overall, setting the default to $k=4$ and using 6 to 8 experts yields robust and efficient performance.

\begin{table}[t]
	\centering
	\caption{Cross-subject accuracy (\%) under different test rate and label ratios, without top-$k$ channel filtering. The best results are in bold.}
	\label{tab:label_ratio}
	\renewcommand{\arraystretch}{0.6}
	\setlength{\tabcolsep}{2pt} 
	\begin{tabularx}{\linewidth}{P{0.08\linewidth} P{0.08\linewidth} *{8}{Y}}
		\toprule
		\multirow{2}{*}{\makecell{Test\\Rate}} &
		\multirow{2}{*}{\makecell{Labeled\\Ratio}} &
		\multicolumn{2}{c}{DEAP} &
		\multicolumn{2}{c}{DREAMER} &
		\multicolumn{4}{c}{SEED} \\
		\cmidrule(lr){3-4}\cmidrule(lr){5-6}\cmidrule(lr){7-10}
		& & V & A & V & A & Inter & S0 & S1 & S2 \\
		\midrule
		~0.1 & ~~0.1 & 70.68 & 66.92 & 72.99 & 85.49 & 66.64 & 71.55 & 69.83 & 73.83 \\
		~0.1 & ~~0.2 & 70.29 & 64.64 & 79.46 & 90.10 & 80.76 & 86.72 & 89.51 & 88.39 \\
		~0.1 & ~~0.3 & 75.01 & 75.92 & 83.91 & 91.15 & 87.96 & \textbf{92.00} & 93.37 & \textbf{93.97} \\
		~0.2 & ~~0.1 & 65.81 & 61.18 & 70.54 & 85.12 & 69.23 & 68.02 & 74.40 & 66.45 \\
		~0.2 & ~~0.2 & 76.20 & 76.86 & 79.58 & 90.96 & 80.56 & 83.65 & 84.65 & 80.64 \\
		~0.2 & ~~0.3 & 82.82 & 85.97 & 85.53 & \textbf{93.38} & \textbf{89.65} & 85.37 & 93.97 & 84.29 \\
		~0.3 & ~~0.1 & 72.87 & 67.47 & 71.57 & 82.40 & 59.99 & 50.16 & 49.55 & 60.12 \\
		~0.3 & ~~0.2 & 82.40 & 82.66 & 78.30 & 88.18 & 77.81 & 54.84 & 90.21 & 72.36 \\
		~0.3 & ~~0.3 & \textbf{87.36} & \textbf{87.46} & \textbf{86.15} & 91.51 & 74.94 & 57.82 & \textbf{95.99} & 77.24 \\
		\bottomrule
	\end{tabularx}
\end{table}

\subsubsection{Effect of test rate and label ratio}

To assess how the test rate and the amount of target labels affect PRISM under the cross-subject setting, we evaluate a grid of configurations without enabling top-$k$ channel filtering. As show in Table \ref{tab:label_ratio}, the results reveal three patterns. \textbf{First}, increasing the label ratio almost always yields gains. With a fixed test rate, raising the labeled ratio from 0.1 to 0.3 leads to improvements on both dimensions of DEAP and DREAMER, and SEED shows concurrent gains for the inter split and all three sessions when the test rate is 0.1 or 0.2. This indicates that more target supervision amplifies the benefits of PRISM. \textbf{Second}, enlarging the test rate is generally unfavorable, most evidently on SEED. At a fixed label ratio, moving the test rate from 0.1 to 0.3 causes systematic drops on S0 and S2, and the inter split also falls when labeled ratio is 0.3, whereas S1 degrades less and can even rise at higher label ratios. This suggests heterogeneous sensitivity of sessions to changes in sample size. \textbf{Third}, robustness differs across datasets. DREAMER on the arousal dimension maintains high performance across settings and increases steadily with more labels. DEAP shows no obvious degradation when the test rate grows and continues to improve with a higher label ratio. Overall, a smaller test rate combined with a larger label ratio is the most reliable regime. When the test rate is large, especially SEED-S0 and SEED-S2, it becomes more sensitive to the specific allocation of data and labels.

\begin{figure*}[htbp]
	\centering{
		\subfigure[{DEAP-V}]{
			\includegraphics[width=0.15\linewidth]{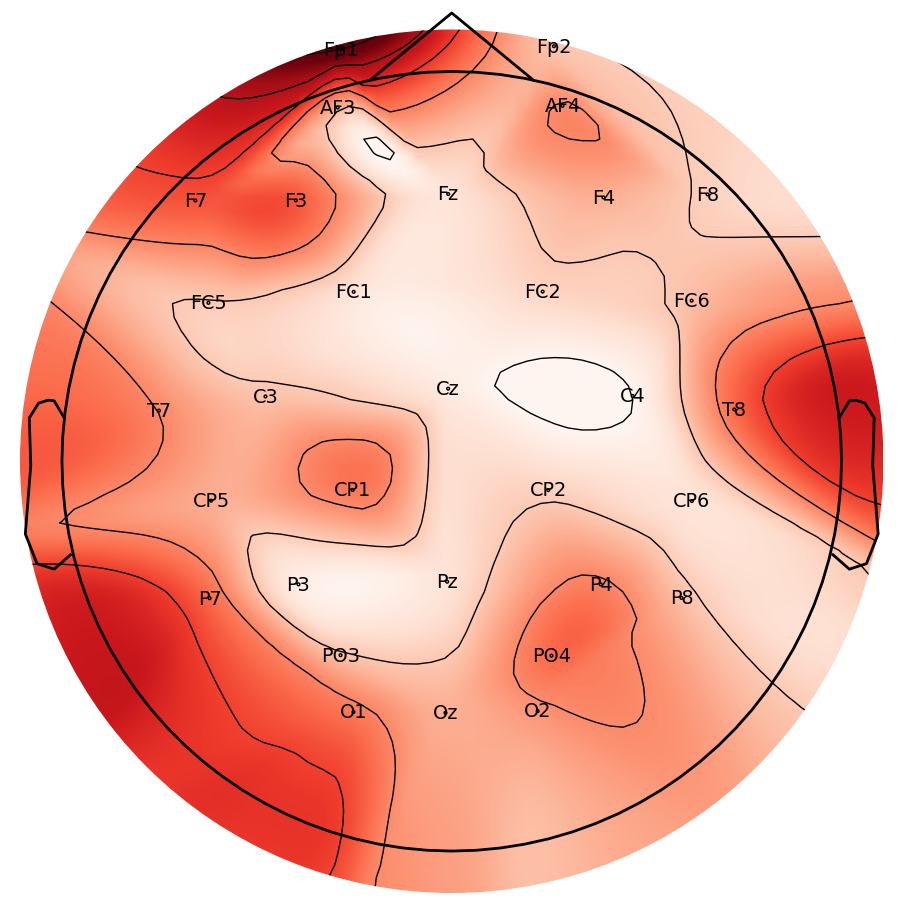}
			\label{fig.deap_v}
		}
		\quad
		\subfigure[{DEAP-A}]{
			\includegraphics[width=0.15\linewidth]{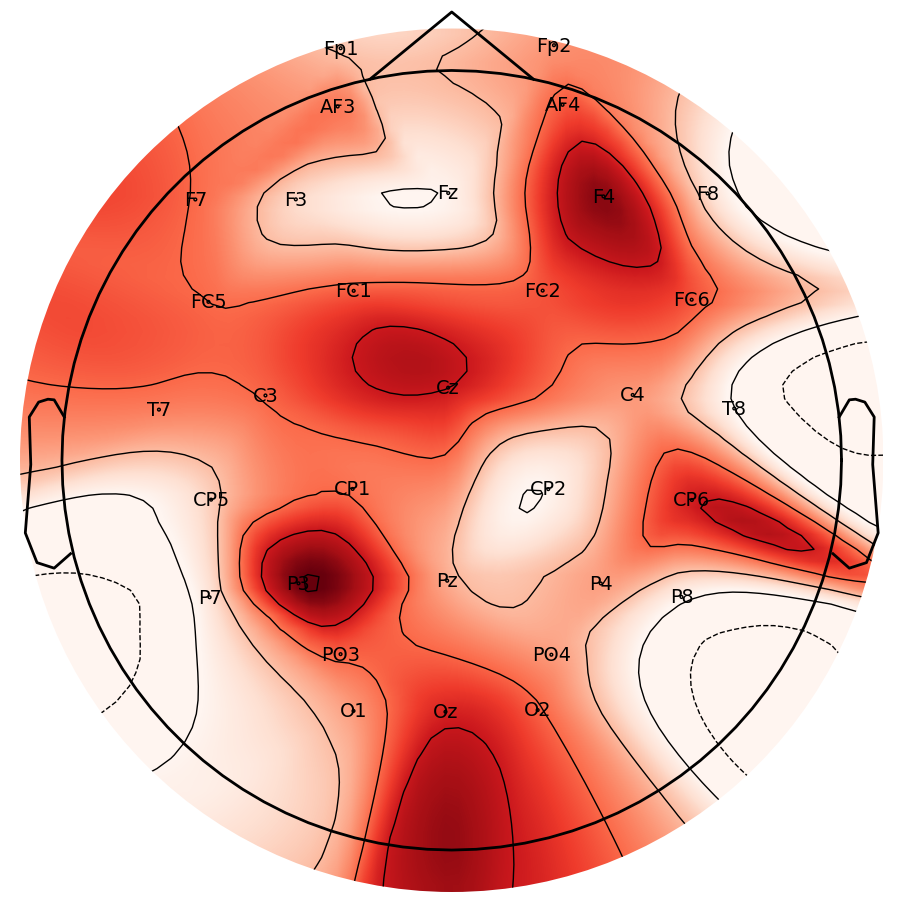}
			\label{fig.deap_a}
		}
		\quad
		\subfigure[{DREAMER-V}]{
			\includegraphics[width=0.175\linewidth]{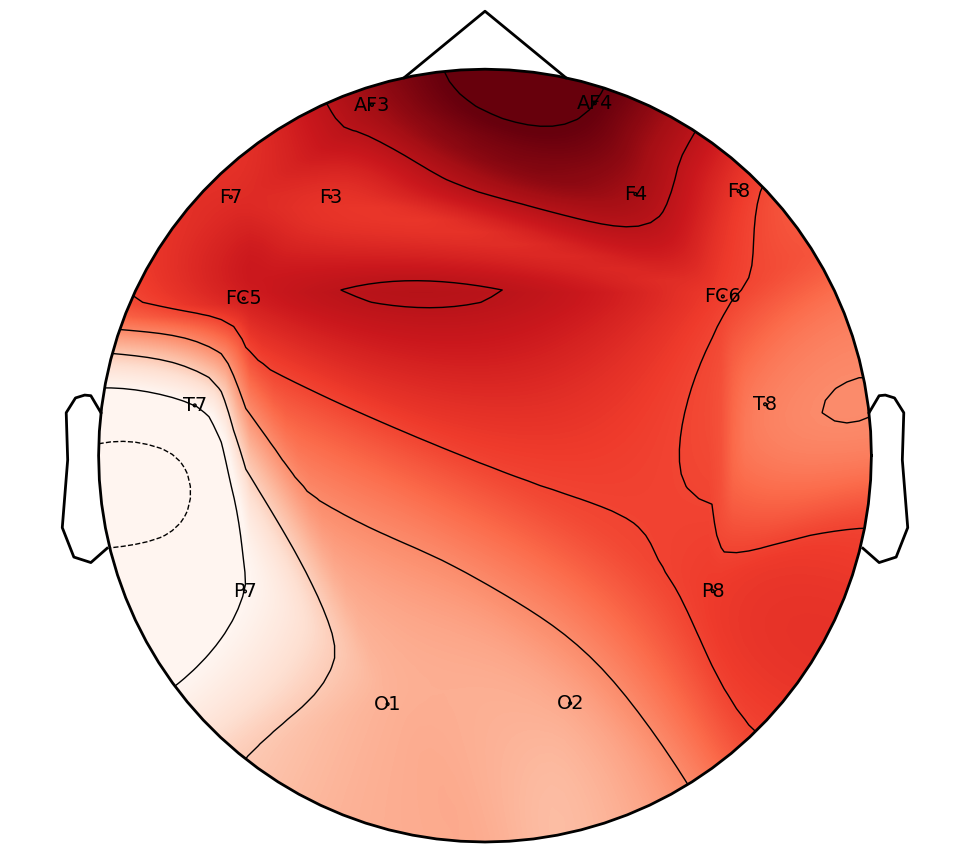}
			\label{fig.dreamer_v}
		}
		\quad
		\subfigure[{DREAMER-A}]{
			\includegraphics[width=0.175\linewidth]{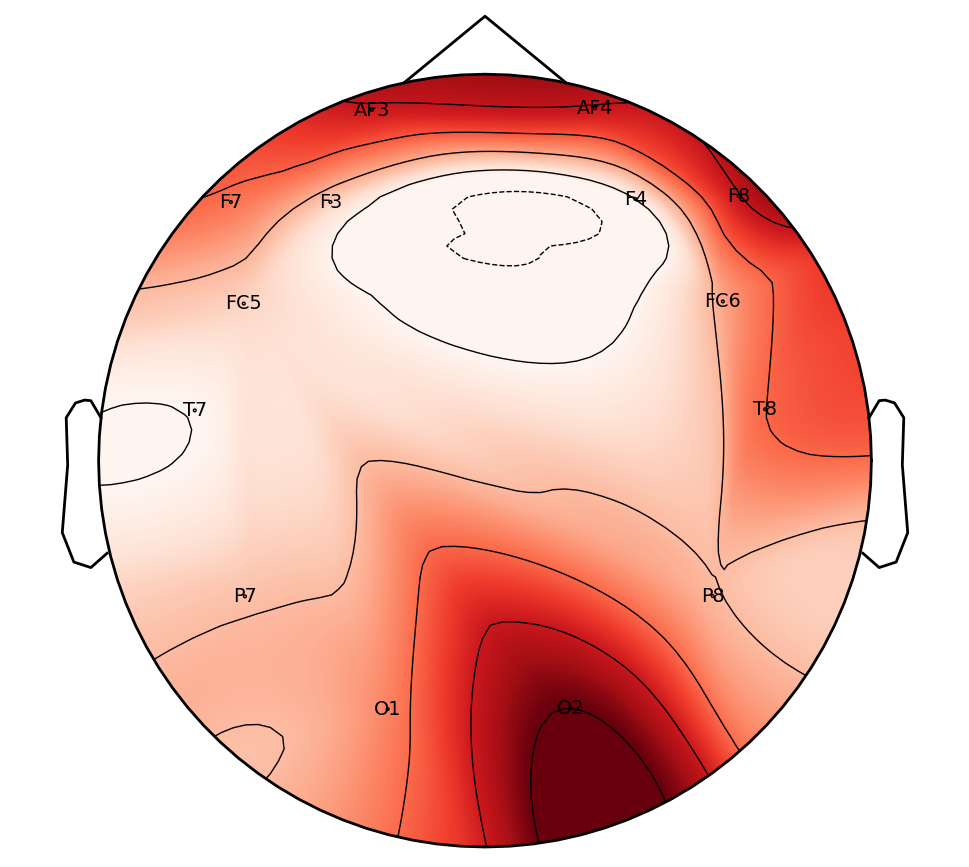}
			\label{fig.dreamer_a}
		}
		\quad
		\subfigure[{SEED}]{
			\includegraphics[width=0.15\linewidth]{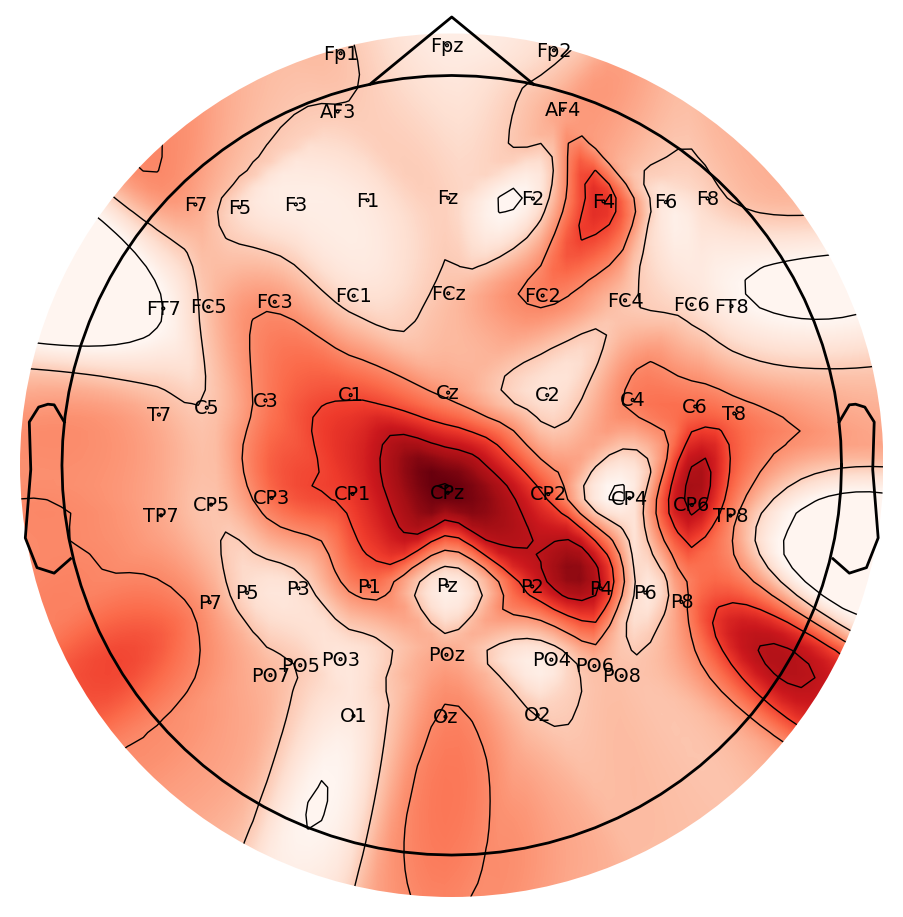}
			\label{fig.seed}
		}
	}
	\caption{PRISM-learned scalp topographies of channel importance across five settings: (a) DEAP–Valence (DEAP-V), (b) DEAP–Arousal (DEAP-A), (c) DREAMER–Valence (DREAMER-V), (d) DREAMER–Arousal (DREAMER-A), and (e) SEED (three-class). Maps show channel weights (normalized for visualization) and warmer colors indicate higher importance. The top-5 channels per setting are: DEAP-V (Fp1, F3, F7, PO4, P7), DEAP-A (P3, F4, Cz, CP6, Oz), DREAMER-V (AF4, AF3, F4, FC5, F7), DREAMER-A (O2, F8, AF4, AF3, T8), and SEED (CPz, CP6, P4, F4, P2).}
	\label{fig.vis}
\end{figure*}

\section{Discussion}\label{sec5}
\subsection{Channel importance analysis}
The core strength of PRISM's Prioritized Channel Importance Module lies in its capacity to dynamically identify and weight the most relevant EEG channels in a data-driven manner, tailored to different datasets and emotional dimensions. To validate this, we visualized the channel importance learned by the model across five experimental settings (DEAP-V/A, DREAMER-V/A, SEED), as presented in Fig. \ref{fig.vis}.

We observed that channels deemed important for valence classification primarily concentrated in the frontal and prefrontal cortices, exhibiting slight hemispheric lateralization. Specifically, DEAP-V was dominated by the left frontal region (Fp1/F3/F7), whereas DREAMER-V showed greater reliance on the bilateral prefrontal area with a slight rightward tendency (AF4/F4). This pattern is consistent with the classical findings linking frontal asymmetry to emotional valence \cite{coan2004frontal}.

In contrast, the channel importance for arousal conditions significantly shifted toward the posterior and right hemispheric regions. High weights in DEAP-A and DREAMER-A concentrated in the occipitotemporal, posterior parietal, and central regions (e.g., O2, T8, CP6, Cz). This spatial distribution aligns with the cognitive control resources accompanying high arousal states \cite{klimesch2012alpha}.

By comparison, the learned weights for the SEED dataset were more distributed, primarily engaging the temporal and parietal regions (e.g., CPz/CP6/P4/F4/P2). We attribute this shift to several factors: the SEED three-classification task (Positive, Negative, Neutral) may dilute the clear frontal asymmetry cues present in pure binary opposition. Furthermore, given SEED’s use of longer video stimuli, enhanced visual processing and emotion-vision interaction are prominent. The known involvement of the right posterior temporal lobe in processing complex visual and emotional cues\cite{vuilleumier2005brains} collectively drives the importance away from the prefrontal cortex toward the parietal and posterior-temporal regions.

\subsection{Analysis of frequency-guided scale selection}

Our multi–scale seasonality mining block selects the top–$k$ frequencies with the highest amplitudes and converts their periods into different scales. A natural concern is that if the selected frequencies concentrate within a narrow band, the resulting different scales may become similar and undermine the goal of multi–scale analysis. To this end, we address this concern from qualitative and quantitative aspects.

\paragraph{Qualitative analysis}
EEG signals typically exhibit activity across multiple frequency bands, including delta (0.5–4 Hz), theta (4–8 Hz), alpha (8–13 Hz), beta (13–30 Hz), and gamma (30–100 Hz). In emotion-related or other cognitively demanding tasks, activity usually appears in more than one band. For example, during an awake state, alpha may dominate under relaxation, whereas theta, beta, and gamma often become pronounced when the cognitive or emotional load increases. This multi–band behavior makes it likely that the top–$k$ frequencies fall into different bands and thus yield diverse scales. As an illustration, at a sampling rate of 128 Hz, selecting frequencies between 8 Hz and 32 Hz produces scales that range approximately from 4 to 16, which is consistent with the intended multi–scale design.

\paragraph{Quantitative evidence on DEAP}
We further quantify the likelihood of frequency concentration using the DEAP dataset. The dataset contains \(N=2{,}457{,}600\) windowed samples. For each sample we identify the top–$k$ frequencies by amplitude in the frequency domain and set \(k=2\). Let \(f_{1i}\) and \(f_{2i}\) denote the two dominant frequencies for the \(i\)-th sample. We compute two statistics:
\begin{equation}
	D \;=\; \frac{1}{N}\sum_{i=1}^{N} \lvert f_{1i}-f_{2i}\rvert ,
\end{equation}
which measures the average absolute distance between the two dominant frequencies, and
\begin{equation}
	R \;=\; \frac{1}{N}\sum_{i=1}^{N} \mathbb{I}\!\left( \lvert f_{1i}-f_{2i}\rvert \le 1 \right),
\end{equation}
which is the proportion of samples whose two dominant frequencies are within 1 Hz, indicating potentially similar scales. On DEAP the empirical results are \(D=17.70\) Hz and \(R=6.05\%\). The average distance indicates a substantial spread between the two dominant frequencies, and the small proportion \(R\) shows that only a small fraction of samples present near identical frequencies. These findings suggest that, in practice, frequency concentration within a narrow band is uncommon for EEG signals and that the risk of degenerate scales is small.

\paragraph{Additional safeguard through MSP}
Even in rare cases where the selected top–$k$ frequencies are close and thus yield similar scales, MSP module preserves multi–scale feature extraction. MSP applies a bank of convolutions with different kernel sizes to the feature maps produced by the temporal block. This design captures local as well as global patterns through diverse receptive fields, maintaining the multi–scale characterization regardless of the exact patch sizes.

Although the concentration of top–$k$ frequencies in a narrow band is a theoretical possibility, qualitative properties of EEG and quantitative evidence on DEAP indicate that this scenario is uncommon. Moreover, the MSP module offers an additional safeguard by enforcing multi–scale receptive fields at the convolutional stage. Overall, these observations support the robustness of our scale selection strategy and the effectiveness of the overall multi–scale design.

\section{Conclusion}\label{sec6}
In this work, we presents a novel framework called PRISM that integrates channel prioritization with semi-supervised domain adaptation. On the modeling side, PRISM emphasizes stable and emotion-relevant electrodes through three coordinated stages, Seasonality Mining, Channelwise State Space, and Expert Router, which capture multi-scale temporal structure, channel dependencies, and channel importance. Under label-scarce target domains, PRISM mitigates cross-subject shift using high-confidence pseudo labels, consistency regularization, and distribution alignment. These components address the dual bottlenecks of channel redundancy and cross-subject distribution shift in EEG emotion recognition. Experiments on DEAP, DREAMER, and SEED across diverse settings demonstrate superior performance. Overall, PRISM shows that jointly modeling channel importance and domain shift is an effective route to improved generalization in EEG emotion recognition, and it offers a plug-and-play solution for label-limited cross-subject applications.



\bibliographystyle{IEEEtran}
\normalem
\bibliography{root}{}

\end{document}